\documentclass[10pt,twocolumn,letterpaper]{article}

\usepackage[pagenumbers]{cvpr} 

\usepackage{graphicx}
\usepackage{amsmath}
\usepackage{amssymb}
\usepackage{booktabs}
\usepackage{kotex}
\usepackage{xcolor}
\usepackage{xspace}
\usepackage{multirow}
\usepackage[T1]{fontenc}
\usepackage{pifont}
\usepackage{comment}
\usepackage{wrapfig}
\usepackage[accsupp]{axessibility}

\usepackage{amsmath,amsfonts,bm}

\def\eqref#1{equation~\ref{#1}}

\def\1{\bm{1}}

\def\vl{{\bm{l}}}

\def\vx{{\bm{x}}}
\def\vy{{\bm{y}}}

\DeclareMathAlphabet{\mathsfit}{\encodingdefault}{\sfdefault}{m}{sl}
\SetMathAlphabet{\mathsfit}{bold}{\encodingdefault}{\sfdefault}{bx}{n}

\def\gD{{\mathcal{D}}}

\def\gS{{\mathcal{S}}}

\def\gX{{\mathcal{X}}}
\def\gY{{\mathcal{Y}}}

\usepackage[pagebackref,breaklinks,colorlinks]{hyperref}

\usepackage[capitalize]{cleveref}
\crefname{section}{Sec.}{Secs.}
\Crefname{section}{Section}{Sections}
\Crefname{table}{Table}{Tables}
\crefname{table}{Tab.}{Tabs.}

\newcommand{\myparagraph}[1]{\vspace{0pt}\noindent{\bf #1}}

\newcommand\oursLLR{$\text{LL-R}$\xspace}
\newcommand\oursLLCt{$\text{LL-Ct}$\xspace}
\newcommand\oursLLCp{$\text{LL-Cp}$\xspace}

\def\cvprPaperID{8402} 
\def\confName{CVPR}
\def\confYear{2022}

\begin{document}

\title{Large Loss Matters in Weakly Supervised Multi-Label Classification}

\author{Youngwook Kim$^{1}\thanks{Equal contribution.}$
~~~~~~~~~
Jae Myung Kim$^{2}$\footnotemark[1]
~~~~~~~~~
Zeynep Akata$^{2,3,4}$
~~~~~~~~~
Jungwoo Lee$^{1,5}$
\\
~\\
\small{
$^1$Seoul National University\quad\quad
$^2$University of T\"{u}bingen\quad\quad}\\
\small{
$^3$Max Planck Institute for Intelligent Systems\quad\quad
$^4$Max Planck Institute for Informatics\quad\quad
$^5$HodooAI Lab}
\vspace{-0.5em}
}
\maketitle

\begin{abstract}

   Weakly supervised multi-label classification (WSML) task, which is to learn a multi-label classification using partially observed labels per image, is becoming increasingly important due to its huge annotation cost. In this work, we first regard unobserved labels as negative labels, casting the WSML task into noisy multi-label classification. From this point of view, we empirically observe that memorization effect, which was first discovered in a noisy multi-class setting, also occurs in a multi-label setting. That is, the model first learns the representation of clean labels, and then starts memorizing noisy labels. Based on this finding, we propose novel methods for WSML which reject or correct the large loss samples to prevent model from memorizing the noisy label. Without heavy and complex components, our proposed methods outperform previous state-of-the-art WSML methods on several partial label settings including Pascal VOC 2012, MS COCO, NUSWIDE, CUB, and OpenImages V3 datasets. Various analysis also show that our methodology actually works well, validating that treating large loss properly matters in a weakly supervised multi-label classification. Our code is available at \url{https://github.com/snucml/LargeLossMatters}.
   
\end{abstract}


\section{Introduction}
\label{sec:intro}

Multi-label classification aims to find all existing objects or attributes in a single image. 
It is gaining attention since the real world is made up of a scene with multiple objects in it \cite{instagram, jft}. Moreover, some of the single-label datasets, also called multi-class datasets, actually have images containing multiple objects \cite{imagenet1, imagenet2}. However, the multi-label classification task has some fundamental difficulties in making a dataset because it requires annotators to label all categories' existence/absence for every image. As the number of categories and images in the dataset increase, annotation cost becomes tremendous \cite{openimages}.

To alleviate these issues, weakly supervised learning approach in multi-label classification task (WSML) has been taken into consideration \cite{wsml1, wsml2, wsml3, prob1}. In a WSML setting, labels are given as a form of partial label, which means only a small amount of categories is annotated per image.
This setting reflects the recently released large-scale multi-label datasets \cite{openimages, lvis} which provide only partial label. Thus, it is becoming increasingly important to develop learning strategies with partial labels. 

\begin{figure}[t]
    \centering
    \includegraphics[width=\linewidth]{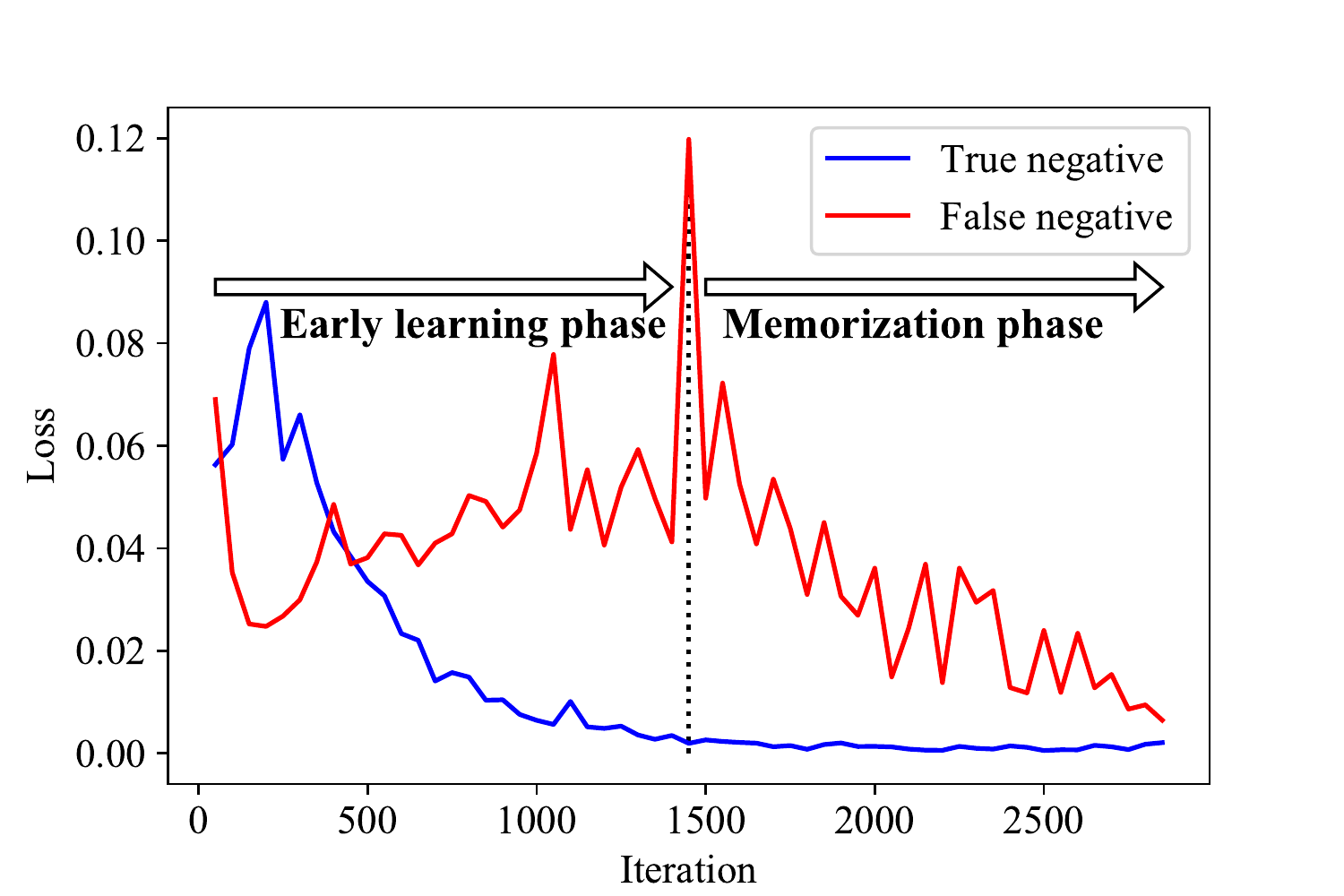}
    \caption{\textbf{Memorization in WSML.} When training ResNet-50 model on PASCAL VOC dataset with partial label, 
    we set all unobserved labels as negative. These labels are composed of true negative and false negative. 
    We observe that the model first fits into true negative label (learning), and then fits into false negative (memorization).
    } 
    \vspace{-10pt}
    \label{fig:memorization}
\end{figure}

There are two naive approaches to train the model with partial labels. One is to train the model with observed labels only, ignoring the unobserved labels. The other is to assume all unobserved labels are negative and incorporate them into training because majorities of labels are negative in a multi-label setting \cite{asl}. 
As the second one has a limitation that this assumption produces some noise in a label which hampers the model learning, previous works \cite{partial_label_2019, imcl, kundu2020exploiting, single_positive_label} mostly follow the first approach and try to explore the cue of unobserved labels using various techniques such as bootstrapping or regularization. However, these approaches include heavy computation or complex optimization pipeline.

We hypothesize that if label noise can be handled properly, the second approach could be a good starting point because it has the advantage of incorporating many true negative labels into model training. Therefore, we try to look at the WSML problem from the perspective of noisy label learning.

Our key observation is about the memorization effect \cite{memorization} in a noisy label learning literature. It is known that when training a model with a noisy label, the model fits into clean labels first and then starts memorizing noisy labels. Although previous work showed the memorization effect only in a noisy \textit{multi-class} classification scenario, we found for the first time that this same effect also happens in a noisy \textit{multi-label} classification scenario.
As shown in Figure \ref{fig:memorization}, during training, the loss value from the clean label (true negative) decreases from the beginning while the loss from the noisy label (false negative) decreases from the middle.



Based on this finding, we borrow the idea from noisy multi-class literature \cite{mentornet, coteaching, lee2019robust} which selectively trains the model with samples having small loss and adapt this idea into a multi-label scenario. Specifically, by assigning the unknown labels as negative in a WSML setting, label noise appears in the form of false negative. Then we develop the three different schemes to prevent false negative labels from being memorized into the multi-label classification model by rejecting or correcting large loss samples during training.

Our method is light and simple, yet effective. It involves negligible computation overhead and does not require complex optimization for model training. Nonetheless, our method surpasses the weakly supervised multi-label classification performance compared to the state-of-the-art methods in Pascal VOC 2012 \cite{pascalvoc}, MS COCO \cite{coco}, NUSWIDE \cite{nuswide}, CUB \cite{cub}, and OpenImages V3 \cite{openimages} datasets. Moreover, while some existing methods are only effective in specific partial label setting \cite{single_positive_label, imcl, partial_label_2019}, our method is broadly applicable in both artificially created and real partial label datasets. Finally, we provide some analysis about the reason why our methods work well from various perspectives.

To sum up, our contributions are as follows;

1) We empirically show for the first time that the memorization effect occurs during noisy multi-label classification.

2) We propose a novel scheme for weakly supervised multi-label classification that explicitly utilizes a learning technique with noisy label.

3) Although light and simple, our proposed method achieves state-of-the-art classification performance on various partial label datasets.

\section{Related Works}

\myparagraph{Multi-label classification.} 
The main research trend of this field has been modeling correlations between labels \cite{multilabelold3, correlation1, correlation2, correlation3} because multiple objects can appear simultaneously in a multi-label setting. Recently this modeling was realized through graph neural networks \cite{gcn1, gcn2, gcn3}, recurrent models \cite{cnnrnn, rnn}, or transformer encoder structure \cite{transformer2021}.
Recent research trends also include solving imbalance issues in multi-label dataset such as long-tail class distribution \cite{longtail1, longtail2} or positive-negative label imbalance \cite{asl}.

\myparagraph{Weakly supervised multi-label classification.} 
Due to the annotation issue, weakly supervised learning of multi-label classification has been another important study. 
There are several approaches to train the model using partially annotated labels: regarding missing labels as negative \cite{wsml1, wsml2, chen2013fast, wang2014binary}, predicting the missing labels via label correlation modeling \cite{wsml3, wsml4, wsml5, deng2014scalable} or probabilistic model \cite{prob1, prob2}. Note that these methods use traditional optimization and they are not scalable to training deep neural networks.

\cite{partial_label_2019} is the first work to train a deep neural network using partial label. It adopts a curriculum learning approach to label some unannotated easy samples using its model prediction. 
However, its initial model trained only on a partial label has a weak representation, which may lead to wrong labelling.
\cite{imcl, kundu2020exploiting} models label similarity and image similarity to predict unobserved labels from other semantically similar images' features or observed labels. 
Recently, \cite{single_positive_label} suggested learning with only one positive label per image, which is a subset of partial label scenario. It also proposed a regularization scheme using an average number of positive labels in a dataset and alternate optimization of classifier and unobserved label estimator.
However, they require complex optimization pipeline or heavy computation cost.
Our method takes a different route with previous method by casting WSML into noisy multi-label classification. Note that few studies have been done in this route except for applying label smoothing \cite{kundu2020exploiting, single_positive_label}.

\myparagraph{Noisy multi-class classification.} In label noise literature, there are two major branches: one is sample selection and the other is label correction.
Sample selection approach starts from the finding of \cite{memorization} and tries to select only clean samples to train the model in the presence of noisy labels. The criterion of clean samples can be small-loss \cite{mentornet, coteaching, lee2019robust, jocor}, consistent prediction with running average of previous predictions \cite{self, elr}, low divergence between prediction and label \cite{josrc}.
Label correction approach tries to update the noisy label instead of viewing it as a fixed one. There are approaches for updating label into softmax-activated prediction \cite{jointopt}, optimizing label via backprop \cite{pencil}, using adaptive target during training \cite{proselflc}. \cite{labelsmoothing-noise} showed that label smoothing can be also viewed as one of the approaches in label correction. There is also a hybrid method \cite{selfie} which takes advantage of both sample selection and label correction. 
Our method borrows the idea of sample selection and label correction to cope with label noise in a WSML setting. However, since the noise type is different between multi-class and multi-label, we propose a method specialized in a multi-label setting.

\section{Approach}
In this section, we start with the definition of assume negative (AN) in weakly supervised multi-label setting (WSML) in \S\ref{sec:assume_negative}. Within this setting, we show in \S\ref{sec:memorization} that the model first learns features of true positive and true negative labels, and then starts memorizing false negative labels. Based on this finding, we propose three methods in \S\ref{sec:method}, that is to modify the large loss samples during training which is likely to be from false negative labels.

\subsection{Target with Assume Negative}
\label{sec:assume_negative}

Let us define an input $\vx \in \gX$ and a target $\vy \in \gY$ where $\gX$ and $\gY$ compose a dataset $\gD$. In a weakly supervised multi-label learning for image classification task, $\gX$ is an image set and $\gY = \{0,1,u\}^K$ where $u$ is an annotation of `unknown', i.e. unobserved label, and $K$ is the number of categories. For the target $\vy$, let $\gS^{\,p}=\{i|y_i=1\}$, $\gS^{\,n}=\{i|y_i=0\}$, and $\gS^{\,u}=\{i|y_i=u\}$. In a partial label setting, small amount of labels are known, thus $|\gS^{\,p}| + |\gS^{\,n}| < K$. We start our method with Assume Negative (AN) where all the unknown labels are regarded as negative. We call this modified target as $\vy^{AN}$,
\begin{equation}
    y^{AN}_i = 
        \begin{cases}
        1, & i \in \gS^{\,p}\\
        0, & i \in \gS^{\,n} \cup  \gS^{\,u}\, ,
        \end{cases}
\end{equation}
and the set of all $\vy^{AN}$ as $\gY^{\,AN}$. $\{y_i^{AN} | i \in \gS^{\,p}\}$ and $\{y_i^{AN} | i \in \gS^{\,n}\}$ are the set where each element is true positive and true negative, respectively. $\{y_i^{AN} | i \in \gS^{\,u}\}$ contains both true negative and false negative.  
The naive way of training the model $f$ with the dataset $\gD^{\,\prime} = (\gX, \gY^{\,AN})$ is to minimize the loss function $L$,
\begin{equation}
    L = \frac{1}{|\gD^{\,\prime}|}
        \sum_{(\vx, \vy^{AN}) \in \gD^{\,\prime}}
        \frac{1}{K} \,
        \sum_{i=1}^{K} \,\, \mathrm{BCELoss} \, (f(\vx)_i, y_i^{AN}) \, , \label{eq:bceloss}
\end{equation}
where $f(\cdot) \in [0,1]^{K}$ and $\mathrm{BCELoss}(\cdot, \cdot)$ is the binary cross entropy loss between the function output and the target. We call this naive method as Naive AN.

\begin{figure*}[t]
    \centering
    \includegraphics[width=\linewidth]{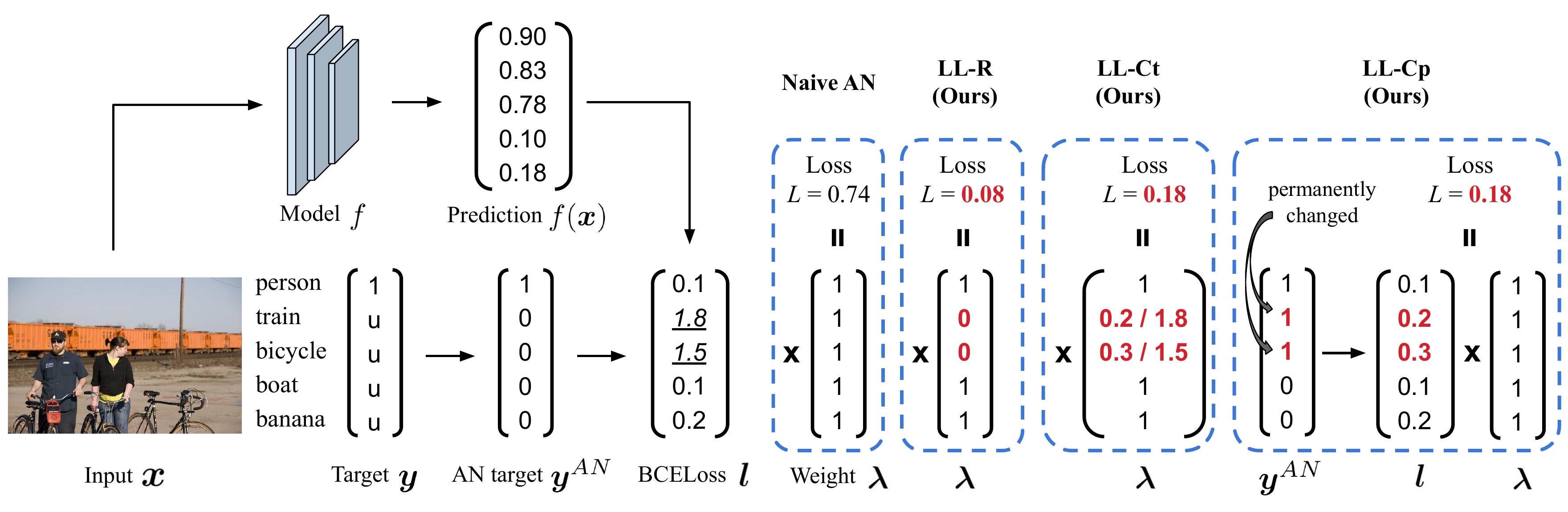}
    \vspace{-15pt}
    \caption{\textbf{Overall pipeline of our proposed methods.} We propose three different ways of dealing false negative labels in AN target $y^{AN}$ which cause large loss. While Naive AN baseline takes average over all elements in $\mathrm{BCELoss}$ $\vl$, our methods control the weight $\lambda$ to reject or correct the false negative labels (\oursLLR or \oursLLCt), or directly change the label from negative to positive (\oursLLCp). Note that 'u' in target $y$ means its label is unobserved.
    } 
    \vspace{-10pt}
    \label{fig:main_figure}
\end{figure*}

\subsection{Memorization in WSML}
\label{sec:memorization}

Let us first revisit the memorization effect in a noisy multi-class learning \cite{memorization}. In the noisy multi-class setting, each data in a dataset is composed of an input and a target where the target is a single category with some of it annotated wrong. For clean labels, the annotated single category is true while for noisy labels the annotated category is false. When a model is trained with the dataset that contains both clean labels and noisy labels, the model first learns features of data with clean labels and then starts to memorize the data with noisy labels. This is in line with the other observation where the model first learns easy patterns and then learns more difficult patterns \cite{deep_image_prior}.

We observe that a similar memorization effect occurs in WSML when the model is trained with the dataset with AN target. To confirm this, we make the following experimental setting. We convert Pascal VOC 2012 \cite{pascalvoc} dataset into partial label one by randomly remaining only one positive label for each image and regard other labels as unknown (dataset $\gD$). These unknown labels are then assumed as negative (dataset $\gD^{\,\prime}$). We train ResNet-50 \cite{resnet} model with $\gD^{\,\prime}$ using the loss function $L$ in Equation \ref{eq:bceloss}. 
We look at the trend of loss value corresponding to each label $y_i^{AN}$ in a training dataset while the model is trained. A single example for true negative label and false negative label is shown in Figure \ref{fig:memorization}. For a true negative label, the corresponding loss value keeps decreasing as the number of iteration increases (blue line). Meanwhile, the loss of a false negative label slightly increases in the initial learning phase, and then reaches the highest in the middle phase followed by decreasing to reach near $0$ at the end (red line). 
This implies that the model starts to memorize the wrong label from the middle phase.

\begin{table}[t]
\centering
\begin{tabular}{c|ccc|ccc}
\multirow{2}{*}{\begin{tabular}[c]{@{}c@{}}Highest loss\\ phase\end{tabular}} & \multicolumn{3}{c|}{Pascal VOC (\%)} & \multicolumn{3}{c}{MS COCO (\%)} \\ \cline{2-7} 
 & TP & TN & FN & TP & TN & FN \\ \hline
Warmup 
& \textbf{88.3} & \textbf{90.7} & 23.8 
& \textbf{64.0} & \textbf{82.6} & 17.3 \\
Regular 
& 11.7 & 9.3 & \textbf{72.2} 
& 36.0 & 17.4 & \textbf{82.7}       
\end{tabular}
\vspace{-5pt}
\caption{\textbf{Distribution of the highest loss occurrence.} For each label, we first draw the loss plot in the training process. We then record whether the highest loss occurred in the warmup phase (epoch 1) or in the regular phase (after epoch 1). TP, TN, FN refers to true positive, true negative, and false negative, respectively.} 
\vspace{-10pt}
\label{tbl:memorization}
\end{table}

To see if this phenomenon constantly occurs across all the labels in a training dataset, we conduct the following experiment.
For every label, we track the loss value on each training epoch.
Then we count the number of labels having the largest loss in the first epoch.
We perform this experiments on partially labeled Pascal VOC 2012 \cite{pascalvoc} and MS COCO dataset \cite{coco} with AN target and ResNet-50. The results are shown in Table \ref{tbl:memorization}.
Most of true positive and true negative samples have a highest loss in the first epoch (warmup phase), whereas false negatives usually show a highest loss after the first epoch (regular phase).
These results indicate that the model learns features from the data corresponding to true positive and true negative labels in the initial phase, while memorization of false negative labels generally starts in the middle of the training phase.

\subsection{Method: Large Loss Modification}
\label{sec:method}

In this section, we propose novel methods for WSML motivated from the ideas of noisy multi-class learning \cite{mentornet, coteaching, lee2019robust} which ignores the large loss during training the model. Remind that in WSML with AN target, 
the model starts memorizing the false negative label in the middle of the training with having a large loss at that time. 
While we can only observe that the label in the set $\{y_i^{AN}| i \in \gS^{\,u}\}$ is negative and cannot explicitly discriminate whether it is false or true, we are able to implicitly distinguish between them.
It is because the loss from false negative is likely to be larger than the loss from true negative before memorization starts. 
Therefore, we manipulate the label in the set $\{y_i^{AN}| i \in \gS^{\,u}\}$ that corresponds to the large loss value during the training process to prevent the model from memorizing false negative labels. We do not manipulate the known true labels, i.e. $\{y_i^{AN}| i \in \gS^{\,p}\cup \gS^{\,n}\}$, since they are all clean labels. Instead of using Equation \ref{eq:bceloss} as a loss function, we further introduce the weight term $\lambda_i$ in the loss function, 
\begin{equation}
    L = \frac{1}{|\gD^{\,\prime}|} 
        \sum_{(\vx, \vy^{AN}) \in \gD^{\,\prime}}
        \frac{1}{K} \,
        \sum_{i=1}^{K} \,\, 
        l_i \times \lambda_i \, . \label{eq:wsml_loss}
\end{equation}
We define $l_i = \mathrm{BCELoss} \, (f(\vx)_i, y_i^{AN})$ where arguments of function $l_i$, that are $f(\vx)$ and $\vy^{AN}$, are omitted for convenience. The term $\lambda_i$ is defined as a function, $\lambda_i=\lambda(f(\vx)_i, y_i^{AN})$, where arguments are also omitted for convenience. $\lambda_i$ is the weighted value for how much the loss $l_i$ should be considered in the loss function $L$ in Equation \ref{eq:wsml_loss}. Intuitively, $\lambda_i$ should be small when $i \in \gS^{\,u}$ and the loss $l_i$ has high value in the middle of the training, that is, to ignore that loss since it is likely to be the loss from a false negative sample. We set $\lambda_i=1$ when $i \in \gS^{\,p}\cup \gS^{\,n}$ since the label $y_i^{AN}$ from these indices is a clean label. We present three different schemes of offering the weight $\lambda_i$ for $i\in\gS^{\,u}$. The schematic description is shown in Figure \ref{fig:main_figure}.

\paragraph{Large loss rejection.} One way of dealing with large loss sample is to reject it by setting $\lambda_i=0$. In a noisy multi-class task, B. Han et al. \cite{coteaching} propose a method of gradually increasing the rejection rate during the training process. We set the function $\lambda_i$ similarly,
\begin{equation}
    \lambda_i = 
        \begin{cases}
        0, & i\in\gS^{\,u} \,\,\,\mathrm{and}\,\,\, l_i > R(t) \\
        1, & \mathrm{otherwise} \, ,
        \end{cases}
\label{eq:ll_relative_rejection}
\end{equation}
where $t$ is the number of current epochs in the training process and $R(t)$ is the loss value that has $[(t-1) \cdot \Delta_{rel}]\%$ largest value in the loss set $\{ l_i \,|\, (\vx, \vy^{AN}) \in \gD^{\,\prime}, i\in\gS^{\,u}\}$. $\Delta_{rel}$ is a hyperparameter that determines the speed of increase of rejection rate. Defining $\lambda_i$ as Equation \ref{eq:ll_relative_rejection} makes rejecting  large loss samples in the loss function $L$. 
We do not reject any loss values at the first epoch, $t=1$, since the model learns clean patterns in the initial phase.
In practice, we use mini-batch in each iteration instead of full batch $D^{\,\prime}$ for composing the loss set. We call this method as \oursLLR.

\paragraph{Large loss correction (temporary).} 
Another way of dealing with large loss sample is correcting rather than rejecting it. 
In a multi-label setting, this can be easily achieved by switching the corresponding annotation from negative to positive.
Specifically, when the loss $l_i$ is large and $i \in \gS^{\,u}$, we \textit{temporarily} modify its label to positive, i.e. $y_i^{AN} = 1$. 
The term ``temporary'' means that it does not change the actual label, but only uses the loss calculated from the modified one.
To reflect this temporary correction scheme in Equation \ref{eq:wsml_loss}, we define the function $\lambda_i$ as
\begin{equation}
    \lambda_i = 
        \begin{cases}
        \frac{\log f(\vx)_i}{\log (1-f(\vx)_i)}, & i\in\gS^{\,u} \,\,\,\mathrm{and}\,\,\, l_i > R(t) \\
        1, & \mathrm{otherwise} \, ,
        \end{cases}
\label{eq:ll_relative_correction}
\end{equation}
where $R(t)$ is same as that in \oursLLR. This makes $l_i \times \lambda_i$ in Equation \ref{eq:wsml_loss} to be the binary cross entropy loss between the function output and positive label when $i\in\gS^{\,u}$ and $l_i > R(t)$ because 
\begin{align}
    l_i \times \lambda_i 
    &= \mathrm{BCELoss}(f(\vx)_i, y_i^{AN}=0) \times \lambda_i \nonumber \\
    &= -\log(1-f(\vx)_i) \times \lambda_i \nonumber\\
    &= - \log f(\vx)_i \nonumber\\
    &= \mathrm{BCELoss}(f(\vx)_i, 1) \, .
\end{align}
We name this method as \oursLLCt. This method has the advantage that it increases the number of true positive labels from unobserved labels.


\paragraph{Large loss correction (permanent).}
In this method, we treat the large loss value more aggressively by \textit{permanently} correcting the label.
We directly change the label from negative to positive and use that modified label from the next training process. To achieve this, we define $\lambda_i = 1$ for every case, and modify the label as follows:
%
\begin{equation}
    y_i^{AN} = 
        \begin{cases}
        1, & i\in\gS^{\,u} \,\,\,\mathrm{and}\,\,\, l_i > R(t) \\
        unchanged, & \mathrm{otherwise} \, ,
        \end{cases}
\label{eq:ll_relative_correction_permanent}
\end{equation}
where $R(t)$ has a constant value of $\Delta_{rel}\%$ largest value in the loss set instead of $[(t-1) \cdot \Delta_{rel}]\%$.   
This makes the number of corrected labels gradually increase as the training progresses.
When the label $y_i^{AN}$ is modified by belonging to the first condition in Equation \ref{eq:ll_relative_correction_permanent}, the set $\gS^{\,u}$ and $\gS^{\,p}$ are also changed as follows:
\begin{align}
    \gS^{\,u} &\leftarrow \gS^{\,u} - \{i\} \, , \\
    \gS^{\,p} &\leftarrow \gS^{\,p} \,\cup \{i\} \, .
\end{align}
%
We name this method as \oursLLCp. 

\paragraph{Absolute variant.}
Instead of gradually increasing the rejection/correction rate, we borrow the idea of using absolute value of loss as a rejection threshold \cite{mentornet} and apply it in WSML. In the rejection and temporary correction schemes, we define the function $\lambda_i$ the same as Equation \ref{eq:ll_relative_rejection} except for $R(t)$ where it is defined as $R(t) = R_0 - t \cdot \Delta_{abs}$. $R_0$ and $\Delta_{abs}$ are hyperparameters where $R_0$ is an initial threshold and $\Delta_{abs}$ determines the speed of decrease of the threshold. 
We report the experimental results of these variant methods in Appendix.

\section{Experiments}
\label{sec:experiments}

In this section, we present experimental results of our method and compare it with previous approaches in two different partial label setting in \S\ref{sec:artificially_experiment} and \S\ref{sec:real_experiment}. In \S\ref{sec:analysis}, we analyze the reason why our methods work well in 5 different ways, that is precision analysis, hyperparameter effect, qualitative results, model explanation, and generalization in a subset of training images.
Throughout this section, we use mean average precision (mAP) as an evaluation metric.

{
\setlength{\tabcolsep}{5pt}
\renewcommand{\arraystretch}{1.2}
\begin{table*}[t]
\centering
\begin{tabular}{c|cccc|cccc}
\multirow{2}{*}{Method} & 
\multicolumn{4}{c|}{End-to-end} & \multicolumn{4}{c}{LinearInit.} \\
&  VOC & COCO & NUSWIDE & CUB  & VOC & COCO & NUSWIDE & CUB \\ \hline\hline
Full label       &    90.2       & 78.0        & 54.5        & 32.9    & 91.1  & 77.2  & 54.9  & 34.0    \\ \hline
Naive AN &  85.1       & 64.1    & 42.0    & 19.1   & 86.9 & 68.7  & 47.6  & 20.9 \\
WAN \cite{mac2019presence, single_positive_label}  &  86.5       & 64.8    & 46.3    & 20.3   & 87.1  & 68.0 & 47.5  & 21.1 \\
LSAN \cite{label_smoothing, single_positive_label}  & 86.7       & 66.9    & 44.9    & 17.9   & 86.5 & 69.2 & 50.5 & 16.6 \\
EPR \cite{single_positive_label}  & 85.5       & 63.3    & 46.0    & 20.0   & 84.9  & 66.8 & 48.1  & 21.2 \\
ROLE \cite{single_positive_label}  &  87.9       & 66.3    & 43.1    & 15.0   & 88.2 & 69.0 & \textbf{51.0} & 16.8 \\  \hline
\oursLLR (Ours)  & \textbf{89.2}       & \textbf{71.0}    & 47.4    & 19.5  & \textbf{89.4}  & \textbf{71.9} & 49.1  & 21.5  \\
\oursLLCt (Ours)  & 89.0  & 70.5  & 48.0 & \textbf{20.4}    & 89.3  & 71.6  & 49.6  & \textbf{21.8} \\
\oursLLCp (Ours) & 88.4   & 70.7  & \textbf{48.3}  & 20.1   & 88.3  & 71.0  & 49.4  &  21.4
\end{tabular}
\vspace{-5pt}
\caption{\textbf{Quantitative results in artificially created partial label datasets.} Results of the model trained with full label are given in the second row to show the upper bound of WSML. ``End-to-end'' indicates that the entire weights of the model is fine-tuned from the beginning, while ``LinearInit.'' indicates the backbone is frozen for the first few epochs. LL-Ct outperforms all baseline methods in 7 out of 8 settings, while LL-R and LL-Cp in 6 out of 8 settings.}
\vspace{-5pt}
\label{tbl:single_positive_label}
\end{table*}
}

\subsection{Artificially created partial label dataset}
\label{sec:artificially_experiment}

\myparagraph{Datasets.} For a multi-label dataset where full labels are annotated, we artificially drop some of its labels for a partial label setting.
Specifically, we follow the procedure presented by \cite{single_positive_label}: for each training image in a dataset, we randomly remain one positive label and regard other labels as unknown.
We experiment on Pascal VOC 2012 \cite{pascalvoc}, MS COCO 2014 \cite{coco}, NUSWIDE \cite{nuswide}, and CUB \cite{cub} datasets. For CUB the task is to classify not the bird categories but the attributes where multiple attributes exist for each image. 

\myparagraph{Implementation details.} For fair comparisons we use the same seed number to create the
same artificial dataset as in \cite{single_positive_label}. We use ResNet-50 \cite{resnet} architecture which is pretrained on ImageNet \cite{krizhevsky2012imagenet} dataset. A single GPU with batch size 16 is used. Each image is resized into 448x448 and performed data augmentation by randomly flipping an image horizontally.
We conduct experiments on two learning schemes. One is using the ``LinearInit'' which first freezes the backbone and update the weights of final linear layer for the initial epochs followed by fine-tuning the entire weights for the remaining epochs, and the other is ``End-to-end'' which is to fine-tune the entire weights from the beginning.
Details about hyperparameter settings are described in Appendix.

\myparagraph{Compared methods.} We compare our method with Naive AN, Weak AN (WAN) \cite{mac2019presence, single_positive_label}, Label Smoothing with AN (LSAN) \cite{label_smoothing, single_positive_label}, EPR \cite{single_positive_label} and ROLE \cite{single_positive_label}. 
Note that some methods using only observed labels without using AN target (Curriculum labeling \cite{partial_label_2019}, IMCL \cite{imcl}) doesn't work in this setting. They give a trivial solution that predicts all labels as positive since only positive labels are observed.

\myparagraph{Results.} As shown in Table \ref{tbl:single_positive_label},
our method is closest to the fully labeled performance, e.g. 1.0 and 6.2 mAP difference in Pascal VOC and MS COCO datasets when fine-tuned end-to-end. Compared with Naive AN and Weak AN which use $\lambda_i = 0$ and $\lambda_i = \frac{1}{K-1}$ when $y_i^{AN}=0$ in Equation \ref{eq:wsml_loss}, respectively, our three different methods all have better performance. 
Our method also surpasses LSAN in all datasets, especially having +4.1 and +2.7 mAP gain on a COCO dataset with End-to-end and LinearInit setting respectively. It implies that our method handles the label noise in AN target better than LSAN.
Moreover, in most datasets, our method also outperforms EPR and ROLE. This result shows that gradually modifying large loss samples helps the model to have better generalization in the presence of false negative labels.

\subsection{Real partial label dataset}
\label{sec:real_experiment}

\myparagraph{Datasets.} To see if our proposed method consistently works on a dataset with real partial label, we use OpenImages V3 \cite{openimages} dataset where there is 3.4M training/42K validation/125K test images with 5,000 classes. In this dataset less than 1\% of labels are annotated. 

\myparagraph{Implementation details.} We use ImageNet-pretrained ResNet-101 architecture and 4 GPUs with batch size 288. Each image is resized into 224x224 and random horizontal flip is applied during training. To better analyze the results, we sort the 5000 categories in ascending order with respect to the number of counted training images and divide them into 5 groups, having 1000 categories for each. Group1 is the group where the number of counted images are the smallest while Group5 is the biggest. We report the mAP results in each group as well as in all groups. 
Details about hyperparameter settings are described in Appendix.


\myparagraph{Compared methods.} We compare our method with Curriculum labeling \cite{partial_label_2019}, and IMCL \cite{imcl}, Naive AN, WAN and LSAN. Naive IU (Ignore Unobserved) is also compared which trains the model only with partial label. Note that ROLE \cite{single_positive_label} do not work because they require storing whole label matrix in a memory which is infeasible.

{
\setlength{\tabcolsep}{5pt}
\renewcommand{\arraystretch}{1.2}
\begin{table}[t]
\centering
\small
\begin{tabular}{c|cccccc}
Method     & G1 & G2 & G3 & G4 & G5 & All Gs \\ \hline\hline
Naive IU & 69.5  & 70.3  & 74.8  & 79.2  & 85.5  & 75.9  \\
Curriculum \cite{partial_label_2019}  & 70.4  & 71.3  & 76.2  & 80.5  & 86.8  & 77.1       \\
IMCL  \cite{imcl}   & 71.0  & 72.6  & 77.6  & 81.8  & 87.3  & 78.1       \\ 
Naive AN   & 77.1  & 78.7  & 81.5  & 84.1  & 88.8   & 82.0 \\
WAN \cite{mac2019presence, single_positive_label} & 71.8 & 72.8 & 76.3 & 79.7 & 84.7 & 77.0 \\
LSAN \cite{label_smoothing, single_positive_label}  & 68.4  & 69.3  & 73.7  & 77.9  & 85.6    & 75.0 \\\hline
\oursLLR (Ours) &    77.4     & 79.1       & 82.0        &    84.5     & 89.5        & 82.5       \\
\oursLLCt (Ours) & 77.7 & 79.3  & 82.1  & 84.7  & 89.4  & \textbf{82.6}  \\
\oursLLCp (Ours) & 77.6  & 79.1   & 81.9 & 84.6   & 89.4   & 82.5   \\

\end{tabular}
\vspace{-5pt}
\caption{\textbf{Quantitative results in OpenImages V3 dataset with real partial label.} 5000 categories are sorted in ascending order with respect to the number of training images in which the label of that category is known and then sequentially grouped from Group1 to Group5 with all groups having the same size. All Gs corresponds to the set of all categories. We observe that \oursLLCt has the best performance, followed by \oursLLCp and \oursLLR.}
\vspace{-5pt}
\label{tbl:partial_label}
\end{table}
}

\begin{figure*}[t]
    \centering
    \includegraphics[width=\linewidth]{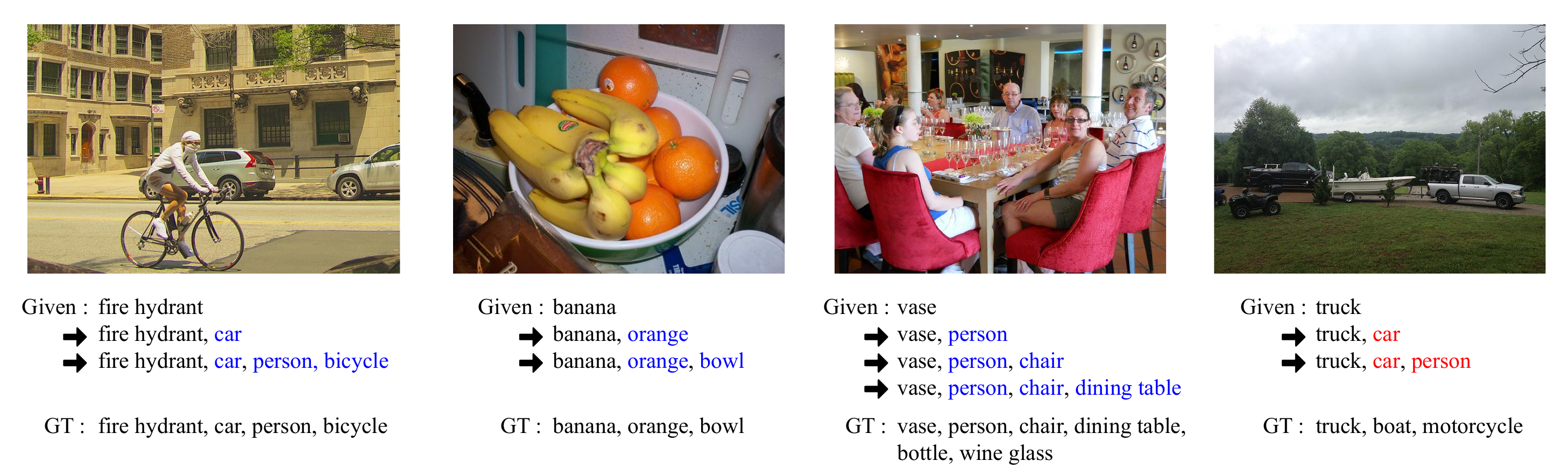}
    \vspace{-15pt}
    \caption{\textbf{Qualitative results in artificially generated COCO partial label dataset.} The arrow indicates the change of categories with positive label during training in our correction scheme \oursLLCt and GT indicates actual ground truth positive labels for a training image. We show three cases where \oursLLCt modifies the unannotated ground truth label correctly, and the failure case at the fourth column.} 
    \vspace{-10pt}
    \label{fig:qualitative}
\end{figure*}
 
\begin{figure}[t]
    \centering
    \includegraphics[width=\linewidth]{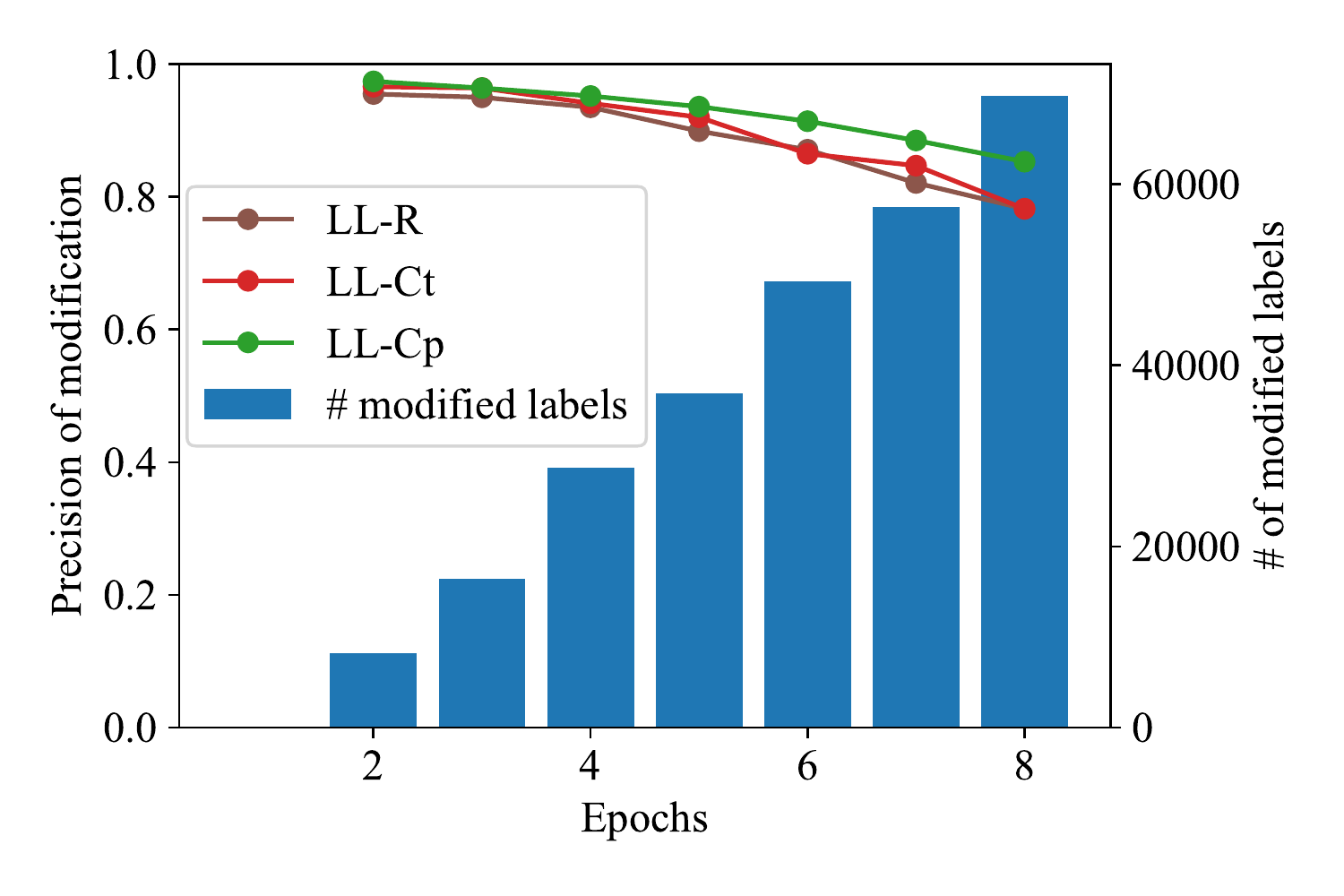}
    \vspace{-25pt}
    \caption{\textbf{Precision analysis of proposed methods on COCO dataset.}} 
    \vspace{-10pt}
    \label{fig:precision}
\end{figure}

\myparagraph{Results.} The results are reported in Table \ref{tbl:partial_label}. 
We first observe that training the model with naive BCE loss with AN target (Naive AN) boosts the classification performance for a large margin compared to previous methods using only the observed labels (Naive IU, Curriculum, IMCL).
We speculate this performance improvement occurred since the average number of observed categories for each image is much smaller than the number of full categories, which hinders the model to be generalized to unseen data when trained with a limited amount of observed labels only. In contrast, even though the AN target is noisy, a large amount of categories may be annotated as true negative after modifying the unobserved labels to the negative labels, making the generalization performance of Naive AN better.

We also observe that \oursLLCt has the best performance of 82.6 mAP, and other methods of ours provide similar high performance. Compared to the Naive AN, our method further rejects or corrects the possible false negative labels, making the degree of noisy labels as less as possible which leads to performance improvement in every groups, from Group1 to Group5. One thing to note is that WAN and LSAN show worse performance than Naive AN, which means that they cannot handle the label noise in AN target in a real partial label scenario.


\subsection{Analysis}
\label{sec:analysis}

In this section, we analyze the reason why our method works well in WSML. Unless mentioned, we perform analysis of our method on an artificially created COCO partial label dataset presented in \S\ref{sec:artificially_experiment} with $\Delta_{rel}=0.2$.

\myparagraph{Precision analysis.} 
To verify whether the label that our proposed methods reject (\oursLLR) or correct (\oursLLCt, \oursLLCp) is actually noisy, we measure the precision of modification. That is, among the labels modified by our scheme as its loss values are large, we calculate the percentage of labels whose actual label is positive. While the precision is calculated in each epoch for \oursLLR and \oursLLCt, we calculate the precision using the accumulated number of labels for \oursLLCp for a fair comparison. We observe in Figure \ref{fig:precision} that our schemes indeed modify the false negative labels with high precision. As the number of epoch increases, precision decreases because the model gradually memorizes the wrong label. 

We can see that \oursLLCp shows the highest precision value among our proposed schemes. However, according to Table \ref{tbl:single_positive_label}, \oursLLCp does not always guarantee highest performance and it may seem a bit contradictory. We conjecture that this is because of the characteristics of \oursLLCp.
As \oursLLCp performs permanent correction, 
erroneously corrected labels may keep damaging the model learning once it is changed.  
Therefore, it might lead to lower mAP even with higher precision of modification.

\begin{figure}[t]
    \centering
    \includegraphics[width=\linewidth]{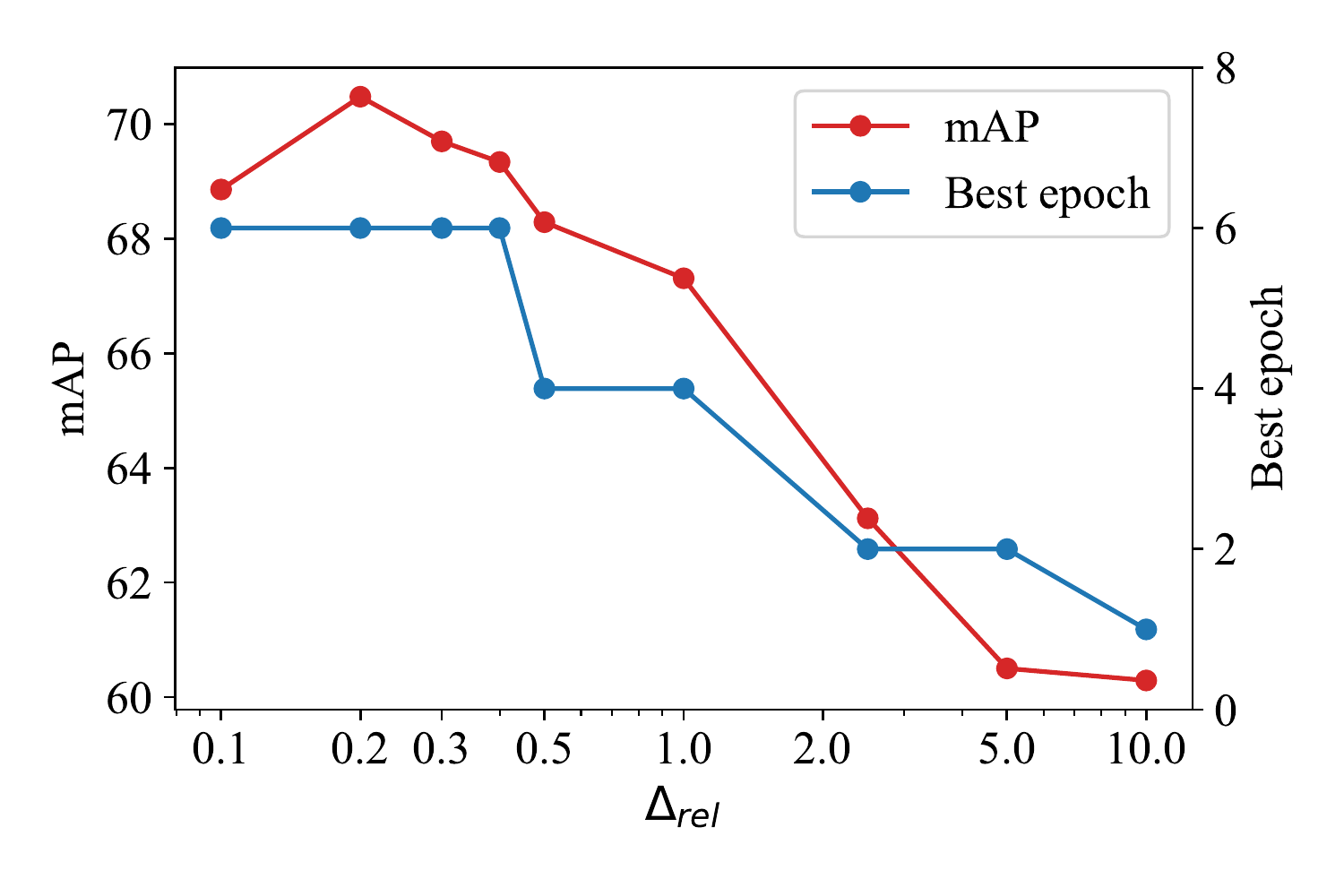}
    \vspace{-25pt}
    \caption{\textbf{Hyperparameter effect of \oursLLCt on COCO dataset.} 
    } 
    \vspace{-10pt}
    \label{fig:hp_coco}
\end{figure}

\myparagraph{Effect of hyperparameter $\Delta_{rel}$.} 
We evaluate the model performance of \oursLLCt with different values of hyperparameter $\Delta_{rel}$ on a COCO dataset. From Figure \ref{fig:hp_coco} we observe that the model produces the best mAP when $\Delta_{rel} = 0.2$. 
When $\Delta_{rel}$ becomes smaller, its performance decreases because the model memorizes false negative labels that are not corrected due to a low correction rate. 
On the other hand, the performance decreases as $\Delta_{rel}$ increases after 0.2. Also, the number of epoch when the model has the best validation score decreases at this time. This is because as $\Delta_{rel}$ increases, our correction scheme wrongly modifies the true negatives labels as positive, making them false positives.
The increased number of false positives hinders the model's generalization, letting model perform early stopping.

\myparagraph{Qualitative results.}
Fig \ref{fig:qualitative} shows the qualitative result of \oursLLCt. The arrow indicates the change of categories with positive labels during training and GT indicates actual ground truth positive labels for a training image. We see that although not all ground truth positive labels are given, our proposed method progressively corrects the category of unannotated GT as positive. We also observe in the first three columns that a category that has been corrected once continues to be corrected in subsequent epochs, even though we perform correction temporarily for each epoch. This conveys that \oursLLCt successfully keeps the model from memorizing false negatives. We also report the failure case of our method on the rightmost side where the model confuses the car as truck which is a similar category and misunderstands the absent category person as present.

\begin{wraptable}{r}{0.45\linewidth}
\centering
\setlength{\tabcolsep}{5pt}
\renewcommand{\arraystretch}{1.2}
\scriptsize
\vspace{-10pt}
\begin{tabular}{c|cc}
Method          & VOC & COCO \\ \hline\hline
Naive AN & 78.9      & 46.4   \\
WAN \cite{single_positive_label, mac2019presence} & 79.8 & 47.7 \\
LSAN \cite{label_smoothing, single_positive_label} & 79.5 & 49.1   \\
EPR \cite{single_positive_label} & 80.2 & 48.1 \\
ROLE \cite{single_positive_label} & 82.5 & 51.5   \\ \hline
\oursLLR (Ours)     & \textbf{83.7}      & 54.0   \\
\oursLLCt (Ours) & \textbf{83.7}  & \textbf{54.1}  \\
\oursLLCp (Ours) & 83.5  & 53.3  \\
\end{tabular}
\vspace{-5pt}
\caption{\textbf{Pointing Game.}}
\vspace{-10pt}
\label{tbl:pointing_game}
\end{wraptable}
%
\myparagraph{Model explanation.} 
We have seen that our methods have quantitatively better performance than other baseline methods. To see if this is related to the model's better understanding of the data, we examine how much the model's explanation is related to the human reasoning process \cite{cam, gradcam, calm}. Concisely, we regard the class activation mapping (CAM) \cite{cam} as the model's explanation and the ground truth object as the human's explanation. To measure how much these two explanations are aligned, we use the Pointing Game metric \cite{pointinggame1, petsiuk2018rise}. For each existing category in an input instance, we consider it as `Hit' if the pixel point of the maximum value in CAM is inside the bounding box of the object, and `Miss' if it is not. We count the \#Hit and \#Miss in all existing categories in all test data, and report the average of \#Hit / (\#Hit + \#Miss) $\times$ 100 calculated for each category in Table \ref{tbl:pointing_game}.

We observe that in both VOC and COCO datasets, our three methods outperform previous methods. In particular, \oursLLCt has +1.2 and +2.6 gain in VOC and COCO datasets compared to ROLE \cite{single_positive_label}, respectively. This result indicates that the explanation of the model trained with our methods is better aligned to human's explanation. We report the CAM visualization results in Appendix.

\begin{figure}[t]
    \centering
    \vspace{-15pt}
    \includegraphics[width=.9\linewidth]{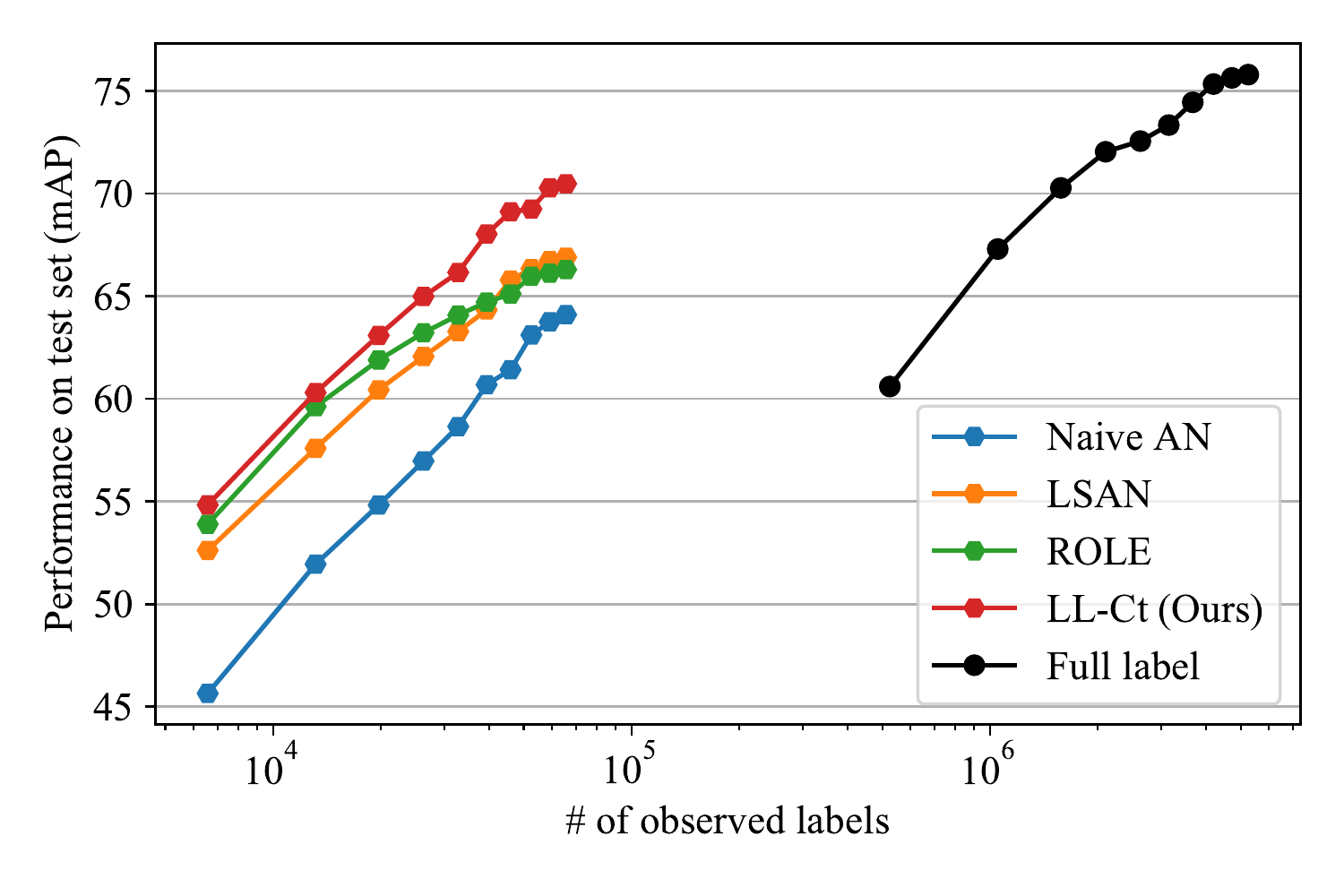}
    \vspace{-10pt}
    \caption{\textbf{Training with smaller number of image.}
    } 
    \vspace{-10pt}
    \label{fig:subset}
\end{figure}

\myparagraph{Training with smaller number of image.}
To see if our method also works in a smaller number of training image, we randomly subsample training images in COCO dataset by 10\%, 20\%, $\cdots$, 90\%, 100\%,  respectively, and train the model with partial label in \S\ref{sec:artificially_experiment} as well as full label.
We then measure the classification performance on test set.

The results are shown in Figure \ref{fig:subset}. While the number of observed labels for weakly supervised methods with 100\% of training image is much more smaller than the fully supervised method with 10\% of training image, i.e. $\times$ 1/8, all the weakly supervised methods outperform the performance with full supervision. Moreover, \oursLLCt showes a similiar performance to the fully supervised method with 30\% training image only with 1/24 of the observed labels. This indicates that when we have a limited cost to annotate the labels when making a multi-label dataset, it is better to weakly annotate many images rather than fully annotate small number of images. We also observe that \oursLLCt outperforms other weakly supervised methods on all ranges of number of observed labels. When only 10\% of training image is given, \oursLLCt has +9.2 mAP better performance compared to the result from Naive AN. This means our method also provides better generalization with small number of training image.

\section{Conclusion}
In this paper, we present large loss modification schemes that reject or correct the large loss samples appearing during training the multi-label classification model with partially labeled annotation. This originates from our empirical observation that memorization effect also happens in a noisy multi-label classification scenario. Although heavy and complex components are not included, our schemes successfully keep the multi-label classification model from memorizing the noisy false negative labels, achieving state-of-the-art performance on various partially labeled multi-label datasets. 

\myparagraph{Limitations and broader impact.}
Since it is difficult to collect enormous data with fully annotated categories, partial label setting is essential \cite{instagram, jft}. For instance, Instagram dataset is composed of billions of social media images with its corresponding hashtags as labels that are used to be noisy \cite{instagram}. Our methodology makes one step progress towards dealing with noisy multi-label classification. However, current WSML methods have limitations that are yet to be reached to the performance with fully annotated label. We hope our methodology facilitates further research in the field of WSML to reach full label performance. \newline

\myparagraph{Acknowledgements.}
This work is in part supported by National Research Foundation of Korea (NRF, 2021R1A4A1030898(10\%)), Institute of Information \& communications Technology Planning \& Evaluation (IITP, 2021-0-00106 (50\%), 2021-0-01059 (20\%), 2021-0-00180 (20\%)) grant funded by the Ministry of Science and ICT (MSIT), Tech Incubator Program for Startups Korea, Ministry of SMEs and Startups, INMAC, and BK21-plus. Also, this work has been partially funded by the ERC (853489 - DEXIM) and by
the DFG (2064/1 – Project number 390727645).

\clearpage

{\small
\bibliographystyle{ieee_fullname}
\bibliography{main}
}

\end{document}


\title{Large Loss Matters in Weakly Supervised Multi-Label Classification \\   - Supplementary Material -}

\author{Youngwook Kim$^{1}\thanks{Equal contribution.}$
~~~~~~~~~
Jae Myung Kim$^{2}$\footnotemark[1]
~~~~~~~~~
Zeynep Akata$^{2,3,4}$
~~~~~~~~~
Jungwoo Lee$^{1,5}$
\\
~\\
\small{
$^1$Seoul National University\quad\quad
$^2$University of T\"{u}bingen\quad\quad}\\
\small{
$^3$Max Planck Institute for Intelligent Systems\quad\quad
$^4$Max Planck Institute for Informatics\quad\quad
$^5$HodooAI Lab}
\vspace{-0.5em}
}
\maketitle

\begin{alphasection}
\section{Hyperparameter settings}
Throughout our experiments, we report the performance of the model with the highest mAP in the validation set.

For \S4.1, we search hyperparameter $\Delta_{rel}$ in $\{0.1, 0.2, 0.3, 0.4, 0.5\}$ and learning rate by dividing the range between values in $\{0.01, 0.001, 0.0001, 0.00001\}$ into quarters. We train the model for 10 epochs in ``End-to-end'' scheme. In ``LinearInit.'' we train the model for 25 epochs with freezing feature extractor and then fine-tune the best model for 10 epochs. Also, we use the same validation set with previous work [7] which use 20\% of training set for validation.

For \S4.2, we train the model for 20 epochs and use the learning rate 2e-5. We search the hyperparameter $\Delta_{rel}$ in \{0.005, 0.01\}. Both for \S4.1 and \S4.2, we set the learning rate of the last fully connected layer as 10 times larger than feature extractor's learning rate.

\section{Experiments for absolute variant}
\label{sec:intro}

In this section, we conduct experiments using the \textit{absolute} value of loss as a large loss criterion, rather than using a \textit{relatively} largest value in a mini-batch. That is, $R(t) = R_0 - t \cdot \Delta_{abs}$ where $R_0$ is an initial threshold and $\Delta_{abs}$ determines the speed of decrease of the threshold. 
Using this criterion, we convert our methods LL-R, LL-Ct, and LL-Cp into LL-R$_{abs}$, LL-Ct$_{abs}$, and LL-Cp$_{abs}$.
We perform hyperparameter search for $R_0$ in $\{1.0, 1.5, 2.0\}$ and $\Delta_{abs}$ in $\{0.1, 0.15, 0.2\}$. We compare the performance of absolute variants with Naive AN and LL-relative, which is the highest performance value among LL-R, LL-Ct, and LL-Cp with the relative threshold.

\myparagraph{Results.}
Table \ref{tbl:single_positive_label_abs} showes the quantitative results in artificially created partial label datasets with fine-tuning the model's entire weight from the beginning. We observe that these absolute variants perform similarly to LL-relative. Especially for NUSWIDE dataset, LL-Ct$_{abs}$ has a +6.6 mAP gain compared to Naive AN, +0.3 mAP gain compared to LL-relative.

Table \ref{tbl:partial_label_abs} represents the quantitative results in OpenImages V3 dataset with real partial label.
Compared to Naive AN, our absolute variants show a +0.3 $\sim$ +0.4 mAP performance gain.

{
\setlength{\tabcolsep}{5pt}
\renewcommand{\arraystretch}{1.2}
\begin{table}[t]
\centering
\begin{tabular}{c|cccc}
\multirow{2}{*}{Method} & 
\multicolumn{4}{c}{End-to-end} \\
&  VOC & COCO & NUSWIDE & CUB   \\ \hline\hline
Naive AN    & 85.1 & 64.1       & 42.0        & 19.1     \\ \hline
LL-relative (Ours) & \textbf{89.2}       & \textbf{71.0}    & 48.3    & \textbf{20.4}    \\ \hline
LL-R$_{abs}$ (Ours) & 89.0       & 70.3    & 47.4    & 20.1   \\
LL-Ct$_{abs}$ (Ours) & 89.0  & 70.3  & \textbf{48.6}  & 19.8   \\
LL-Cp$_{abs}$ (Ours) & 88.7   & 70.5  & 48.0  & 19.8   
\end{tabular}
\vspace{-5pt}
\caption{\textbf{Quantitative results in artificially created partial label datasets.} }
\vspace{-15pt}
\label{tbl:single_positive_label_abs}
\end{table}
}

{
\setlength{\tabcolsep}{5pt}
\renewcommand{\arraystretch}{1.2}
\begin{table}[t]
\centering
\small
\begin{tabular}{c|cccccc}
Method     & G1 & G2 & G3 & G4 & G5 & All Gs \\ \hline\hline
Naive AN   & 77.1  & 78.7  & 81.5  & 84.1  & 88.8   & 82.0 \\ \hline
LL-relative (Ours) & 77.7 & 79.3  & 82.1  & 84.7  & 89.4  & \textbf{82.6}  \\ \hline

LL-R$_{abs}$ (Ours)  & 77.9       & 78.8       & 81.7       & 84.1       & 88.9 & 82.3 \\
LL-Ct$_{abs}$ (Ours) & 77.5 & 79.1  & 81.8  & 84.4  & 89.3  & 82.4  \\
LL-Cp$_{abs}$ (Ours) & 77.7  & 79.0   & 81.7 & 84.3   & 88.9   & 82.3  
\end{tabular}
\vspace{-5pt}
\caption{\textbf{Quantitative results in OpenImages V3 dataset with real partial label.} }
\label{tbl:partial_label_abs}
\end{table}
}


\section{Model Explanation}
In this section, we visualize the Class Activation Map (CAM) result of ground truth categories for some of the COCO dataset test images. We compare the mapping results with Naive AN and ROLE. As shown in Figure \ref{fig:cam}, our LL-Ct has the capability of capturing the location of an object more than previous methods.
This indicates that our method's explanation is better aligned to human's explanation.

\begin{figure*}[t]
    \centering
    \includegraphics[width=\linewidth]{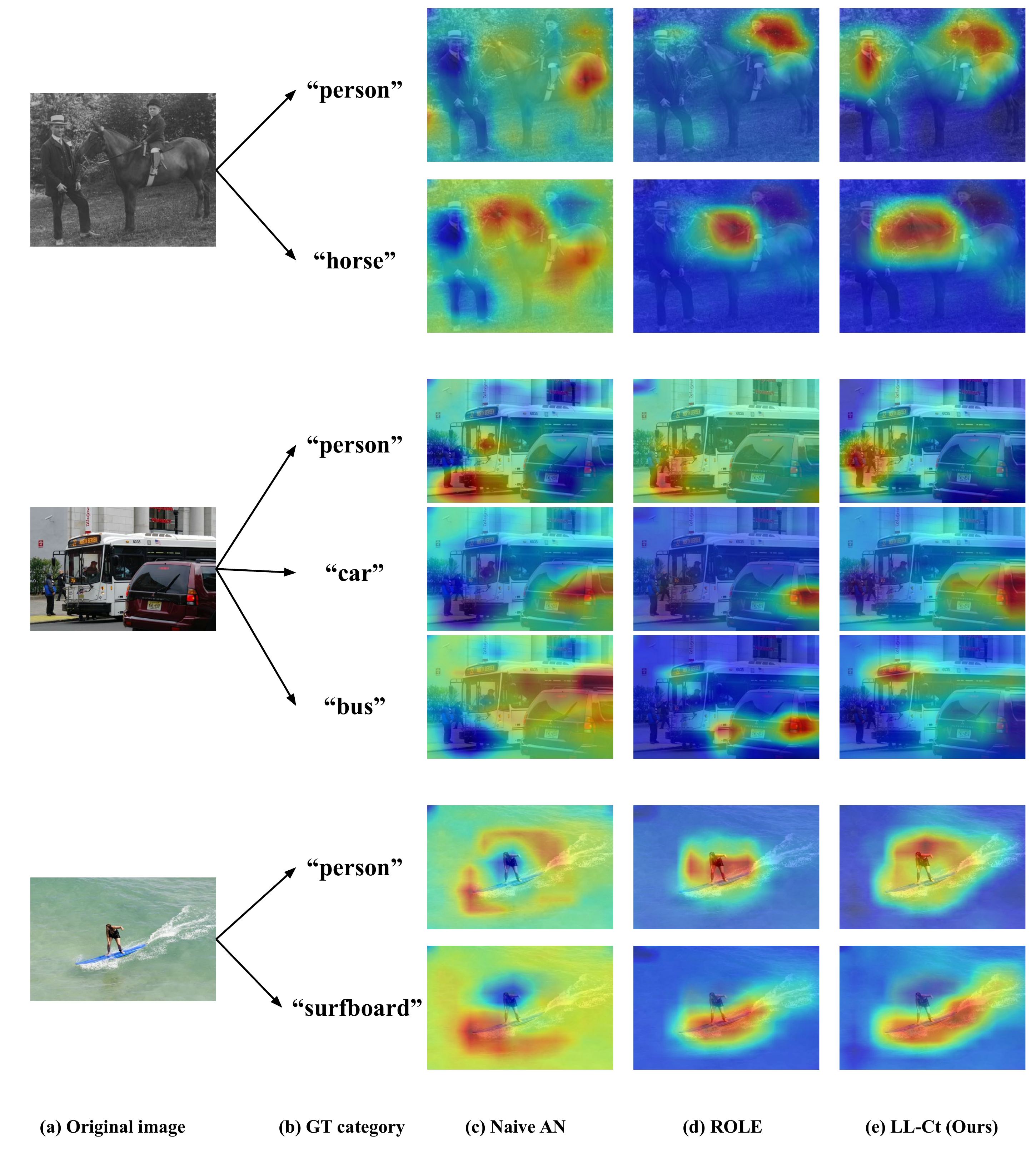}
    \vspace{-15pt}
    \caption{\textbf{Class Activation Mapping results in COCO test images.}} 
    \vspace{-10pt}
    \label{fig:cam}
\end{figure*}
\end{alphasection}